\documentclass{article}

\usepackage[nonatbib,final,preprint]{neurips_2020}

\usepackage[utf8]{inputenc} % allow utf-8 input
\usepackage[T1]{fontenc}    % use 8-bit T1 fonts
\usepackage{hyperref}       % hyperlinks
\usepackage{url}            % simple URL typesetting
\usepackage{booktabs}       % professional-quality tables
\usepackage{amsfonts}       % blackboard math symbols
\usepackage{nicefrac}       % compact symbols for 1/2, etc.
\usepackage{microtype}      % microtypography

\usepackage[pdftex]{graphicx}
\usepackage{subcaption}
\usepackage{float} 
\graphicspath{{images/}}

\usepackage{mathrsfs}
\newcommand{\Lagrangian}{\mathscr{L}}
\newcommand{\Difficulty}{D}

\newcommand{\Action}{\mathscr{A}}

\title{Unsupervised Difficulty Estimation\\with Action Scores}

\author{%
  Octavio Arriaga\\
  Department of Computer Science\\
  University of Bremen\\
  28359 Bremen, Germany \\
  \texttt{arriagac@uni-bremen.de} \\
  \And
  Matias Valdenegro-Toro \\
  Robotics Innovation Center\\
  German Research Center for Artificial Intelligence\\
  28359 Bremen, Germany \\
  \texttt{matias.valdenegro@dfki.de}
}

\begin{document}
\maketitle

\begin{abstract}
    Evaluating difficulty and biases in machine learning models has become of extreme importance as current models are now being applied in real-world situations.
    In this paper we present a simple method for calculating a difficulty score based on
    the accumulation of losses for each sample during training. We call this the action score. Our proposed method does not require any modification of the model neither any external supervision, as it can be implemented as callback that gathers information from the training process.
    We test and analyze our approach in two different settings: image classification, and object detection, and we show that in both settings the action score can provide insights about model and dataset biases.
\end{abstract}
    
\begin{figure}[H]
    \centering
    \begin{subfigure}[b]{0.13\textwidth}
        \includegraphics[width=\textwidth]{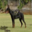}
        \caption{\scriptsize{Dog 1015.9}}
        \label{fig:cifar10_hard_a}
    \end{subfigure}
    \begin{subfigure}[b]{0.13\textwidth}
        \includegraphics[width=\textwidth]{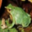}
        \caption{\scriptsize{Cat 958.6}}
        \label{fig:cifar10_hard_b}
    \end{subfigure}
    \begin{subfigure}[b]{0.13\textwidth}
        \includegraphics[width=\textwidth]{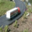}
        \caption{\scriptsize{Truck 854.4}}
    \end{subfigure}
    \begin{subfigure}[b]{0.13\textwidth}
        \includegraphics[width=\textwidth]{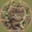}
        \caption{\scriptsize{Cat 893.2}}
    \end{subfigure}
    \begin{subfigure}[b]{0.13\textwidth}
        \includegraphics[width=\textwidth]{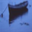}
        \caption{\scriptsize{Ship 776.8}}
    \end{subfigure}
    \begin{subfigure}[b]{0.13\textwidth}
        \includegraphics[width=\textwidth]{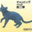}
        \caption{\scriptsize{Cat 752.8}}
    \end{subfigure}
    \begin{subfigure}[b]{0.13\textwidth}
        \includegraphics[width=\textwidth]{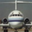}
        \caption{\scriptsize{Plane 743.9}}
    \end{subfigure}\label{fig:cifar10_hard}

    \centering
    \begin{subfigure}[b]{0.13\textwidth}
        \includegraphics[width=\textwidth]{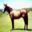}
        \caption{\scriptsize{Horse 0.073}}
    \end{subfigure}
    \begin{subfigure}[b]{0.13\textwidth}
        \includegraphics[width=\textwidth]{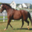}
        \caption{\scriptsize{Horse 0.0280}}
    \end{subfigure}
    \begin{subfigure}[b]{0.13\textwidth}
        \includegraphics[width=\textwidth]{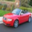}
        \caption{\scriptsize{Car 0.288}}
    \end{subfigure}
    \begin{subfigure}[b]{0.13\textwidth}
        \includegraphics[width=\textwidth]{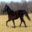}
        \caption{\scriptsize{Horse 0.291}}
    \end{subfigure}
    \begin{subfigure}[b]{0.13\textwidth}
        \includegraphics[width=\textwidth]{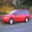}
        \caption{\scriptsize{Car 0.305}}
    \end{subfigure}
    \begin{subfigure}[b]{0.13\textwidth}
        \includegraphics[width=\textwidth]{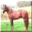}
        \caption{\scriptsize{Horse 0.322}}
    \end{subfigure}
    \begin{subfigure}[b]{0.13\textwidth}
        \includegraphics[width=\textwidth]{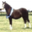}
        \caption{\scriptsize{Horse 0.335}}
    \end{subfigure}\caption{Most difficult (top-row) and easiest examples (bottom-row) in CIFAR10. Our proposed \textit{action score} is displayed below each image as well as the true label.}
    \label{fig:cifar10ActionScores}

\end{figure}

\section{Introduction}
    Current state-of-the-art models in computer vision tasks rely on the use of convolutional neural networks (CNNs).
    However, modern CNN architectures contain sufficient structural-priors to reduce the solution space to a computable and generalisable one, but not restricted enough to prevent them from learning unstructured data nuances~\cite{Zhang2016, nguyen2015deep, jo2017measuring, goodfellow2014explaining}.
    In this paper we present a simple method to assess the difficulty and possible biases of machine learning models by tracking the loss of each sample during training.
    This method does not rely in any external supervision nor model modification as opposed to similar methods~\cite{shrivastava2016training, loshchilov2015online, lin2017focal, wang2018towards}.
    Specifically, we test it in a simple image classification scenario and a more complex setting with a multi-objective loss used in object detection.
    
    The use of per-sample loss values is widespread in the literature.	
    \cite{shrivastava2016training} uses the per-sample loss to mine for hard negative examples while training an object detector.
    \cite{loshchilov2015online} proposes a way to sample mini-batches using the loss as a criteria, where training samples with higher loss will be chosen more frequently.
    This has the effect of speeding up training by 5$\times$.
    The focal loss \cite{lin2017focal} introduces a similar concept where an object detector focuses on harder samples.    
    Difficulty estimation is an emerging topic in this field.
    \cite{wang2018towards} proposes an additional output branch and a related loss function in order to learn to estimate sample difficulty.
    This method has learning difficulties and cannot be trained end-to-end.    
    
\section{Unsupervised Difficulty Estimation}
    Given a loss function $\Lagrangian$ and a model $m$ with free parameters $\theta_{n}$, we define the action $\Action$ 
    of a sample $x \in X$ with labels $y \in Y$ as
    \begin{equation}
        \Action(x) = \sum_{n=0}^{N} \Lagrangian(y, m(x; \theta_n))
    \end{equation}
    where $n$ represents epochs.
    Consequently, the action\footnote{We adopt this name due to its similarity of a physical system following the path of stationary action~\cite{Landau1960}.}
    of a sample is the accumulated loss over all epochs.
    Our method characterizes the action of each sample as a measurement of its difficulty.
    Therefore, samples with a higher accumulated loss represent samples that are more difficult to learn.
    Specifically, we argue that the action is directly proportional to its difficulty i.e. $\Difficulty(x) \propto \Action(x)$.
    Within this framework we can also recover sample pairs that accumulate the least amount of loss during optimization.
    These samples reflect which elements are easier to learn as well as possible biases that might be present in the data.
    We would like to emphasize that the method presented here can be applied to any learning algorithm that is optimized iteratively and is not limited to artificial neural networks nor supervised methods.

\section{Results}
    We first tested our method in simple classification task in which we train a VGG-like CNN\footnote{We used the Keras CIFAR10 example CNN available at \href{https://github.com/keras-team/keras/blob/master/examples/cifar10_cnn.py}{keras-examples}} on CIFAR10 using the cross-entropy loss.
    At every epoch we calculated and stored the loss of each sample in the test set.
    After the conclusion of the training phase we calculated the action of each sample by summing up the stored losses.
    In Figure~\ref{fig:cifar10ActionScores} we display the samples with the most and least action scores.
    From Figure~\ref{fig:cifar10ActionScores} we can observe that model learns to distinguish with the least action two specific set of samples: brown horses and red cars.
    For our second experiment we calculate the action scores of a multi-objective loss function used for training the single-shot object detector SSD300~\cite{Liu2016}.
    The total loss of this model consist of the combination of three different losses: positive classification, negative classification and bounding box regression. 
    For the localization loss the samples with the most and least action are shown in Figure~\ref{fig:pascalLocalActionScores}   
    
    We can observe that the most difficult samples for the box regression loss correspond to images that contain undistinguishable small objects.
    Moreover, easier samples for the same loss are determined by single centered objects.
    % We can observe in Figures \ref{fig:pascal_positive_hard_a} and \ref{fig:pascal_positive_hard_b} that the object-classes are presented in non-conventional situations.
    % Figures \ref{fig:pascal_positive_hard_e}, \ref{fig:pascal_positive_hard_f} and \ref{fig:pascal_positive_hard_g} show objects that complicated to identify.
    % In Figure \ref{fig:pascal_positive_easy} we observe that persons are cats are the two objects classes that are easier for the model to learn.
    % \input{pascal_hard_positive_loss.tex}
    % \input{pascal_easy_positive_loss.tex}
    % Figures \ref{fig:pascal_negative_hard} and \ref{fig:pascal_negative_easy} display the most and least actions scores for negative classification.
    % Specifically in Figure \ref{fig:pascal_negative_hard} we find that most of the samples are cluttered environments.
    % Consequently, it becomes difficult for the model to correctly classify negative boxes.
    % Similar to Figure \ref{fig:pascal_positive_easy}, Figure \ref{fig:pascal_negative_easy} contains images of single centered objects.
    % \input{pascal_hard_negative_loss.tex}
 
    We provide additional examples of object detection on PASCAL VOC 2007 in the supplementary material.
\section{Conclusions and Future Work}
    In this work we presented a method for calculating the difficultly and possible biases of a model.
    Our method requires no external supervision nor a modification of the original model and it can be easily integrated in any learning framework.
    We test our method in two different settings.
    We displayed the samples with the highest and lowest \textit{actions scores}.
    % We calculated the accumulated losses for each sample in the test set and displayed those samples with the highest and lowest values. 
    Our obtained results indicate that the maximum and minimum \textit{action scores} do qualitatively correspond to difficult or biased samples.
    For future work we propose to apply our method in unsupervised settings, as well as to test its variability along different models. 

% \section*{References}
\clearpage
\bibliographystyle{plain}
\bibliography{references}

\clearpage
\appendix

\section{Object Detection Results on PASCAL VOC 2007 with SSD}

In this section we show results on the PASCAL VOC 2007 validation set using the Single Shot Multibox detector \cite{Liu2016}. SSD uses a multi-task loss, a localization loss for bounding box regression, and a cross-entropy loss for class predictions. The cross-entropy loss can be divided into loss for the positive examples (target objects), and loss for the negative examples (background). We show results in each components of the multi-task loss, namely localization, positive, and negative losses.

\begin{figure}[H]
    \centering
    \begin{subfigure}[b]{.13\textwidth}
        \includegraphics[width=\textwidth]{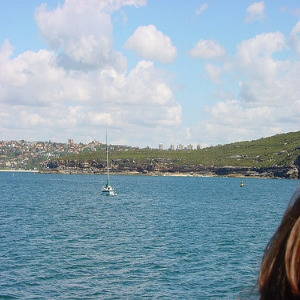}
        \caption{\scriptsize{Boat 1749.3}}
    \end{subfigure}
    \begin{subfigure}[b]{.13\textwidth}
        \includegraphics[width=\textwidth]{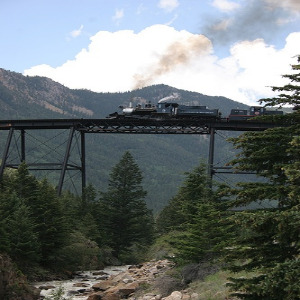}
        \caption{\scriptsize{Train 496.6}}
    \end{subfigure}
    \begin{subfigure}[b]{.13\textwidth}
        \includegraphics[width=\textwidth]{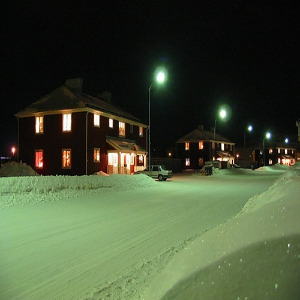}
        \caption{\scriptsize{Car 457.1}}
    \end{subfigure}
    \begin{subfigure}[b]{.13\textwidth}
        \includegraphics[width=\textwidth]{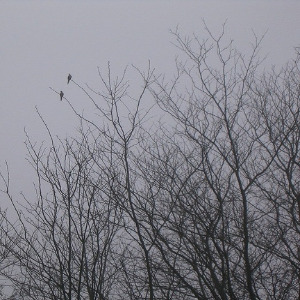}
        \caption{\scriptsize{Bird 413.1}}
    \end{subfigure}
    \begin{subfigure}[b]{.13\textwidth}
        \includegraphics[width=\textwidth]{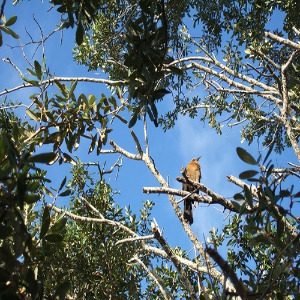}
        \caption{\scriptsize{Bird 385.8}}
    \end{subfigure}
    \begin{subfigure}[b]{.13\textwidth}
        \includegraphics[width=\textwidth]{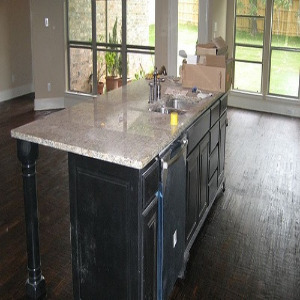}
        \caption{\scriptsize{Plant 357.6}}
    \end{subfigure}
    \begin{subfigure}[b]{.13\textwidth}
        \includegraphics[width=\textwidth]{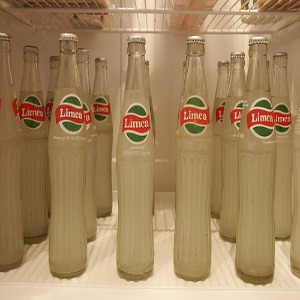}
        \caption{\scriptsize{Bottle 354.5}}
    \end{subfigure}

    \centering
    \begin{subfigure}[b]{0.13\textwidth}
        \includegraphics[width=\textwidth]{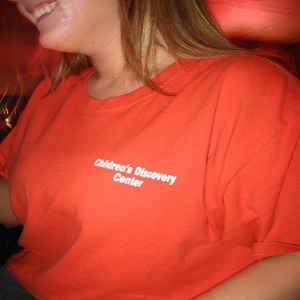}
        \caption{\scriptsize{Person 3.9}}
    \end{subfigure}
    \begin{subfigure}[b]{0.13\textwidth}
        \includegraphics[width=\textwidth]{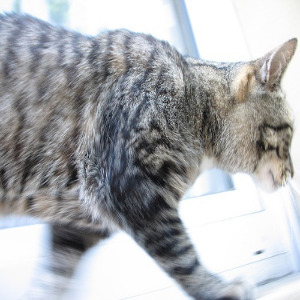}
        \caption{\scriptsize{Cat 4.1}}
    \end{subfigure}
    \begin{subfigure}[b]{0.13\textwidth}
        \includegraphics[width=\textwidth]{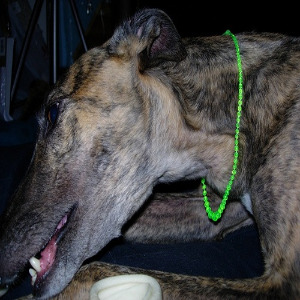}
        \caption{\scriptsize{Dog 4.2}}
    \end{subfigure}
    \begin{subfigure}[b]{0.13\textwidth}
        \includegraphics[width=\textwidth]{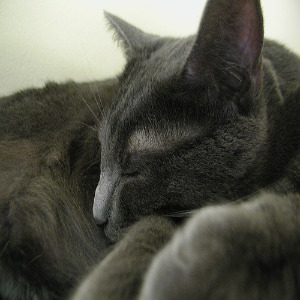}
        \caption{\scriptsize{Cat 4.7}}
    \end{subfigure}
    \begin{subfigure}[b]{0.13\textwidth}
        \includegraphics[width=\textwidth]{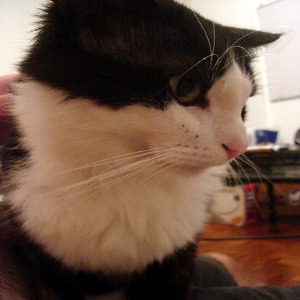}
        \caption{\scriptsize{Cat 5.0}}
    \end{subfigure}
    \begin{subfigure}[b]{0.13\textwidth}
        \includegraphics[width=\textwidth]{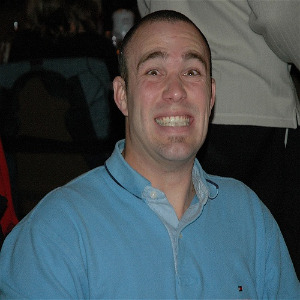}
        \caption{\scriptsize{Person 5.2}}
    \end{subfigure}
    \begin{subfigure}[b]{0.13\textwidth}
        \includegraphics[width=\textwidth]{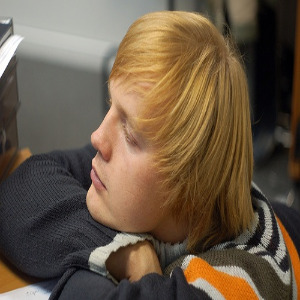}
        \caption{\scriptsize{Person 5.3}}
    \end{subfigure}

    \caption{Most difficult (top-row) and easiest examples (bottom-row) in the VOC 2007-VAL with the SSD localization loss. The \textit{action scores} are displayed below each image as well as the true label.}\label{fig:pascalLocalActionScores}
\end{figure}

\begin{figure}[H]
    \centering
    \begin{subfigure}[b]{0.13\textwidth}
        \includegraphics[width=\textwidth]{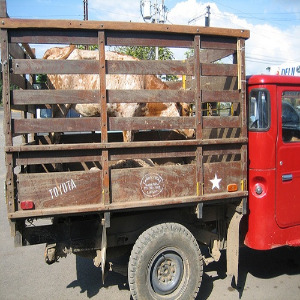}
        \caption{Cow\\596.6}
        \label{fig:pascal_positive_hard_a}
    \end{subfigure}
    \begin{subfigure}[b]{0.13\textwidth}
        \includegraphics[width=\textwidth]{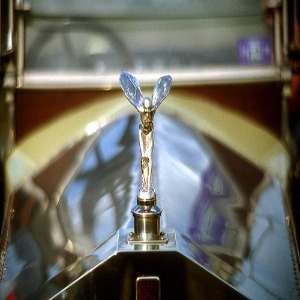}
        \caption{Car\\568.5}
        \label{fig:pascal_positive_hard_b}
    \end{subfigure}
    \begin{subfigure}[b]{0.13\textwidth}
        \includegraphics[width=\textwidth]{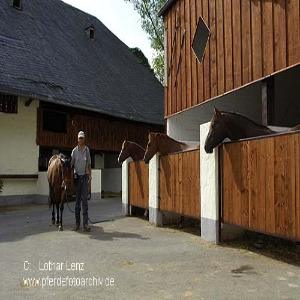}
        \caption{Person and Horse 565.0}
        \label{fig:pascal_positive_hard_c}
    \end{subfigure}
    \begin{subfigure}[b]{0.13\textwidth}
        \includegraphics[width=\textwidth]{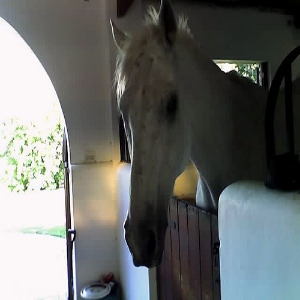}
        \caption{Horse\\544.6}
        \label{fig:pascal_positive_hard_d}
    \end{subfigure}
    \begin{subfigure}[b]{0.13\textwidth}
        \includegraphics[width=\textwidth]{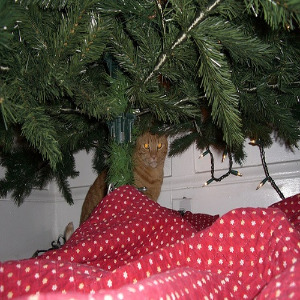}
        \caption{Cat\\544.3}
        \label{fig:pascal_positive_hard_e}
    \end{subfigure}
    \begin{subfigure}[b]{0.13\textwidth}
        \includegraphics[width=\textwidth]{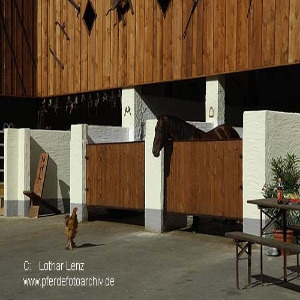}
        \caption{\scriptsize{Plant, Bottle and Horse 527.8}}
        \label{fig:pascal_positive_hard_f}
    \end{subfigure}
    \begin{subfigure}[b]{0.13\textwidth}
        \includegraphics[width=\textwidth]{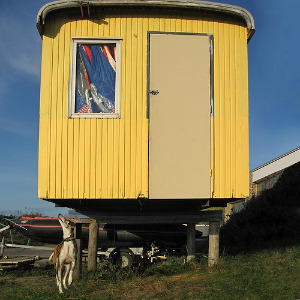}
        \caption{Dog\\524.7}
        \label{fig:pascal_positive_hard_g}
    \end{subfigure}

    % \begin{subfigure}[b]{0.13\textwidth}
        % \includegraphics[width=\textwidth]{pascal-voc/pascal-voc-SSD-hardest-val_positive-7-train-506.117.jpg}
        % \caption{Train\\506.1}
    % \end{subfigure}
    % \begin{subfigure}[b]{0.13\textwidth}
        % \includegraphics[width=\textwidth]{pascal-voc/pascal-voc-SSD-hardest-val_positive-8-sofa-chair-491.331.jpg}
        % \caption{Sofa and Chair\\491.3}
    % \end{subfigure}

    \caption{Hardest Examples on PASCAL VOC 2007 with SSD validation positive loss. Action score is included in each caption.}
    \label{fig:pascal_positive_hard}
\end{figure}

\begin{figure}[H]
    \centering
    \begin{subfigure}[b]{0.13\textwidth}
        \includegraphics[width=\textwidth]{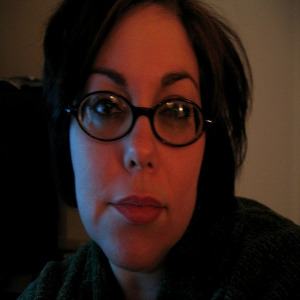}
        \caption{Person\\5.7}
    \end{subfigure}
    \begin{subfigure}[b]{0.13\textwidth}
        \includegraphics[width=\textwidth]{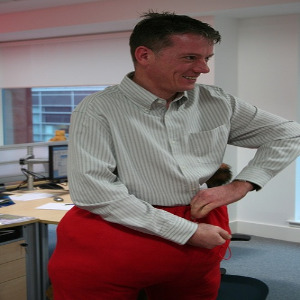}
        \caption{Person\\6.1}
    \end{subfigure}
    \begin{subfigure}[b]{0.13\textwidth}
        \includegraphics[width=\textwidth]{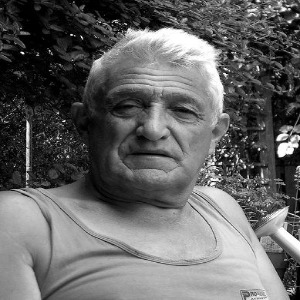}
        \caption{Person\\6.2}
    \end{subfigure}
    \begin{subfigure}[b]{0.13\textwidth}
        \includegraphics[width=\textwidth]{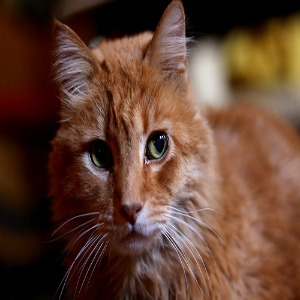}
        \caption{Cat\\6.2}
    \end{subfigure}
    \begin{subfigure}[b]{0.13\textwidth}
        \includegraphics[width=\textwidth]{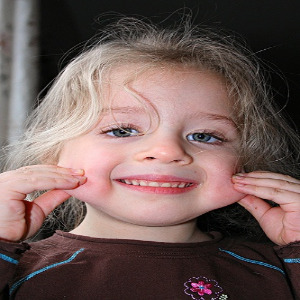}
        \caption{Person\\6.3}
    \end{subfigure}
    \begin{subfigure}[b]{0.13\textwidth}
        \includegraphics[width=\textwidth]{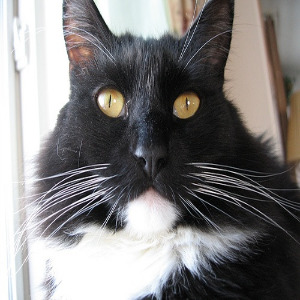}
        \caption{Cat\\6.4}
    \end{subfigure}
    \begin{subfigure}[b]{0.13\textwidth}
        \includegraphics[width=\textwidth]{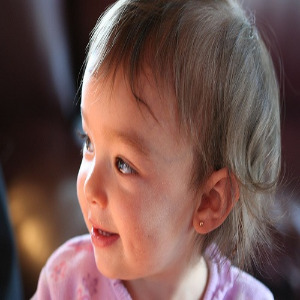}
        \caption{Person\\6.7}
    \end{subfigure}

    % \begin{subfigure}[b]{0.13\textwidth}
        % \includegraphics[width=\textwidth]{pascal-voc/pascal-voc-SSD-easiest-val_positive-7-person-6.769.jpg}
        % \caption{Person\\6.8}
    % \end{subfigure}
    % \begin{subfigure}[b]{0.13\textwidth}
        % \includegraphics[width=\textwidth]{pascal-voc/pascal-voc-SSD-easiest-val_positive-8-cat-6.829.jpg}
        % \caption{Cat\\6.8}
    % \end{subfigure}

    \caption{Easiest Examples on PASCAL VOC 2007 with SSD validation positive loss. Action score is included in each caption.}
    \label{fig:pascal_positive_easy}
\end{figure}

\begin{figure}[H]
    \centering
    \begin{subfigure}[b]{0.13\textwidth}
        \includegraphics[width=\textwidth]{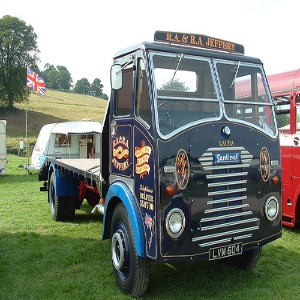}
        \caption{Person\\1086.9}
    \end{subfigure}
    \begin{subfigure}[b]{0.13\textwidth}
        \includegraphics[width=\textwidth]{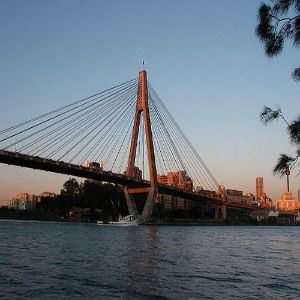}
        \caption{Boat\\626.9}
    \end{subfigure}
    \begin{subfigure}[b]{0.13\textwidth}
        \includegraphics[width=\textwidth]{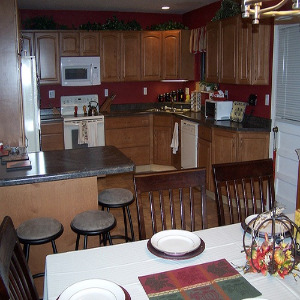}
        \caption{Table\\548.1}
    \end{subfigure}
    \begin{subfigure}[b]{0.13\textwidth}
        \includegraphics[width=\textwidth]{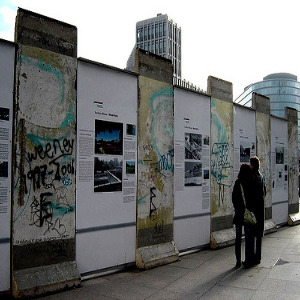}
        \caption{Person\\476.6}
    \end{subfigure}
    \begin{subfigure}[b]{0.13\textwidth}
        \includegraphics[width=\textwidth]{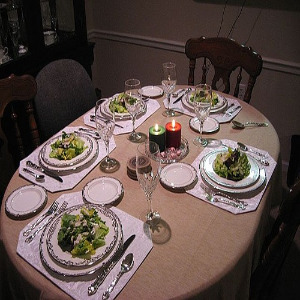}
        \caption{Table\\452.9}
    \end{subfigure}
    \begin{subfigure}[b]{0.13\textwidth}
        \includegraphics[width=\textwidth]{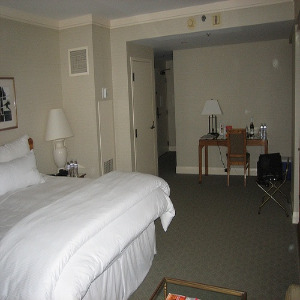}
        \caption{Chair\\416.2}
    \end{subfigure}
    \begin{subfigure}[b]{0.13\textwidth}
        \includegraphics[width=\textwidth]{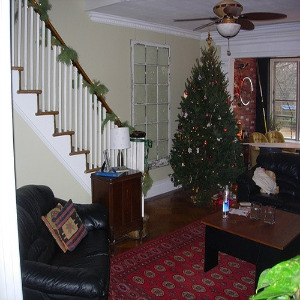}
        \caption{Bicycle\\401.5}
    \end{subfigure}

    % \begin{subfigure}[b]{0.13\textwidth}
        % \includegraphics[width=\textwidth]{pascal-voc/pascal-voc-SSD-hardest-val_negative-7-tvmonitor-391.012.jpg}
        % \caption{TV Monitor\\391.0}
    % \end{subfigure}
    % \begin{subfigure}[b]{0.13\textwidth}
        % \includegraphics[width=\textwidth]{pascal-voc/pascal-voc-SSD-hardest-val_negative-8-bottle-385.261.jpg}
        % \caption{Bottle\\385.3}
    % \end{subfigure}

    \caption{Hardest Examples on PASCAL VOC 2007 with SSD validation negative loss. Action score is included in each caption.}
    \label{fig:pascal_negative_hard}
\end{figure}

\begin{figure}[H]
    \centering
    \begin{subfigure}[b]{0.13\textwidth}
        \includegraphics[width=\textwidth]{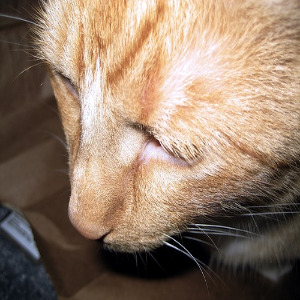}
        \caption{Cat\\6.5}
    \end{subfigure}
    \begin{subfigure}[b]{0.13\textwidth}
        \includegraphics[width=\textwidth]{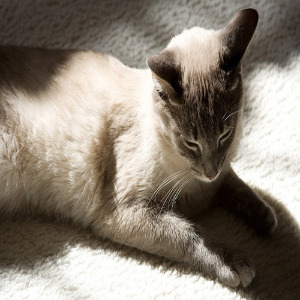}
        \caption{Cat\\8.2}
    \end{subfigure}
    \begin{subfigure}[b]{0.13\textwidth}
        \includegraphics[width=\textwidth]{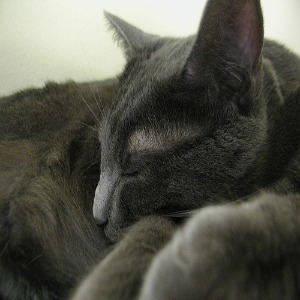}
        \caption{Cat\\8.5}
    \end{subfigure}
    \begin{subfigure}[b]{0.13\textwidth}
        \includegraphics[width=\textwidth]{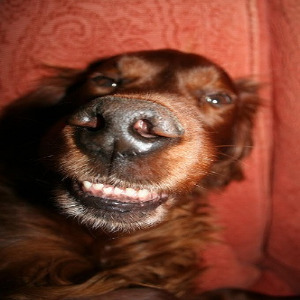}
        \caption{Dog\\8.6}
    \end{subfigure}
    \begin{subfigure}[b]{0.13\textwidth}
        \includegraphics[width=\textwidth]{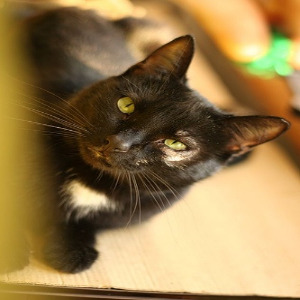}
        \caption{Cat\\8.7}
    \end{subfigure}
    \begin{subfigure}[b]{0.13\textwidth}
        \includegraphics[width=\textwidth]{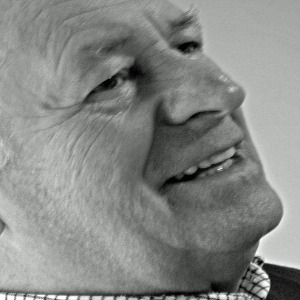}
        \caption{Person\\9.6}
    \end{subfigure}
    \begin{subfigure}[b]{0.13\textwidth}
        \includegraphics[width=\textwidth]{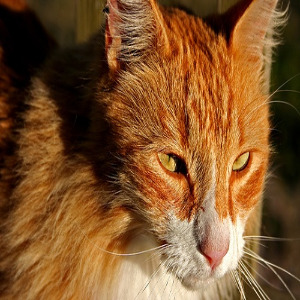}
        \caption{Cat\\10.0}
    \end{subfigure}

    % \begin{subfigure}[b]{0.13\textwidth}
        % \includegraphics[width=\textwidth]{pascal-voc/pascal-voc-SSD-easiest-val_negative-7-person-10.112.jpg}
        % \caption{Person\\10.1}
    % \end{subfigure}
    % \begin{subfigure}[b]{0.13\textwidth}
        % \includegraphics[width=\textwidth]{pascal-voc/pascal-voc-SSD-easiest-val_negative-8-cat-11.594.jpg}
        % \caption{Cat\\6.8}
    % \end{subfigure}

    \caption{Easiest Examples on PASCAL VOC 2007 with SSD validation negative loss. Action score is included in each caption.}
    \label{fig:pascal_negative_easy}
\end{figure}

\end{document}